\documentclass[journal]{IEEEtran}
\usepackage[linesnumbered,ruled]{algorithm2e}
\usepackage{graphicx,amssymb,mathrsfs,amsmath,array,color}
\usepackage{subfigure}
\usepackage{multirow}
\usepackage{booktabs}

\newcommand{\norm}[1]{\lVert#1\rVert}
\DeclareMathOperator{\tr}{tr}
\DeclareMathOperator{\F}{F}
\DeclareMathOperator{\T}{T}

\newcommand{\maxIter}{\textit{maxIter}}

\begin{document}
\title{Learning Shared Cross-modality Representation Using Multispectral-LiDAR and Hyperspectral Data}

\author{Danfeng Hong,~\IEEEmembership{Student Member,~IEEE,}
        Jocelyn Chanussot,~\IEEEmembership{Fellow,~IEEE,}
        Naoto Yokoya,~\IEEEmembership{Member,~IEEE,}
        Jian Kang,~\IEEEmembership{Student Member,~IEEE,} 
        and~Xiao Xiang Zhu,~\IEEEmembership{Senior Member,~IEEE}
        
\thanks{This work is jointly supported by the German Research Foundation (DFG) under grant ZH 498/7-2, the Helmholtz Association under the framework of the Young Investigators Group SiPEO (VH-NG-1018), and the European Research Council (ERC) under the European Unions Horizon 2020 research and innovation programme (grant agreement No. ERC-2016-StG-714087, Acronym: So2Sat). This work of N. Yokoya is also supported by the Japan Society for the Promotion of Science (KAKENHI 18K18067).}
\thanks{D. Hong and X. Zhu are with the Remote Sensing Technology Institute (IMF), German Aerospace Center (DLR), 82234 Wessling, Germany, and Signal Processing in Earth Observation (SiPEO), Technical University of Munich (TUM), 80333 Munich, Germany. (e-mail: danfeng.hong@dlr.de; xiaoxiang.zhu@dlr.de)}
\thanks{J. Chanussot is with the Univ. Grenoble Alpes, Inria, CNRS, Grenoble INP, LJK, F-38000 Grenoble, France, also with the Faculty of Electrical and Computer Engineering, University of Iceland, Reykjavik 101, Iceland. (e-mail: jocelyn@hi.is)}
\thanks{N. Yokoya is with the Geoinformatics Unit, RIKEN Center for Advanced Intelligence Project (AIP), RIKEN, 103-0027 Tokyo, Japan. (e-mail: naoto.yokoya@riken.jp)}
\thanks{J. Kang is with the Signal Processing in Earth Observation (SiPEO), Technical University of Munich (TUM), 80333 Munich, Germany. (e-mail: jian.kang@tum.de)}
}

\markboth{Special Stream of IEEE Geoscience and Remote Sensing Letters,~Vol.~XX, No.~XX, ~XXXX,~2019}
{Shell \MakeLowercase{\textit{et al.}}: }
\maketitle

\begin{abstract}
Due to the ever-growing diversity of the data source, multi-modality feature learning has attracted more and more attention. However, most of these methods are designed by jointly learning feature representation from multi-modalities that exist in both training and test sets, yet they are less investigated in absence of certain modality in the test phase. To this end, in this letter, we propose to learn a shared feature space across multi-modalities in the training process. By this way, the out-of-sample from any of multi-modalities can be directly projected onto the learned space for a more effective cross-modality representation. More significantly, the shared space is regarded as a latent subspace in our proposed method, which connects the original multi-modal samples with label information to further improve the feature discrimination. Experiments are conducted on the multispectral-Lidar and hyperspectral dataset provided by the 2018 IEEE GRSS Data Fusion Contest to demonstrate the effectiveness and superiority of the proposed method in comparison with several popular baselines.
\end{abstract}


\section{Introduction}
\IEEEPARstart{R}{emote} sensing (RS) is one of the most common ways to extract relevant information about Earth and our environment. RS acquisitions can be done by both active (synthetic aperture radar, LiDAR) and passive (optical and thermal range, multispectral and hyperspectral) devices. The complementary of the data acquired by different platforms can be helpful for more accurately characterizing land use \cite{dalla2015challenges}. In this letter we will focus on the joint use of multispectral-Lidar (MS-Lidar) data, providing detailed information about the ground elevation, and hyperspectral image (HSI), providing information of the physical nature of the sensed materials, using a general cross-modality learning (CML) framework \cite{hong2019cospace}.

Intuitively, most existing multi-modality feature learning methods basically follow the concatenation-based fusion strategy \cite{yokoya2018open}. However, either early fusion or latter one might be incapable of effectively addressing the aforementioned challenge, as there is a lack of completely-paired multi-modality samples in the whole dataset. The problem setting naturally motivates us to find a latent shared feature space by learning modality-specific projections from the training samples. 

For this purpose, some tentative works have been proposed by joint dimensionality reduction or alignment learning. In \cite{hong2019cospace}, principal component analysis (PCA) is used to simultaneously project multi-modal data into a common subspace. Rasti \textit{et al.} \cite{rasti2017hyperspectral} fused the HSI and Lidar data using total variation component analysis for land-cover and land-use mapping. Hong \textit{et al.} \cite{hong2017learning} jointly embedded the spatial-spectral information for HSI classification. Besides, manifold alignment (MA) has been proven to be another powerful solution. Following it, Tuia \emph{et al.} \cite{tuia2014ManifoldAlignment} proposed to align multi-view RS imagery on manifolds to reduce the gap between multi-modalities. More generally, Banerjee \textit{et al.} \cite{banerjee2016domain} transferred the samples of source and target domains into a shared latent domain where the learned features in both domains are expected to be consistent. Du \textit{et al.} \cite{du2018multi} for the first time took multi-modal RS data analysis as an unsupervised multi-task learning problem, and proposed a state-of-the-art and milestone blind source separation algorithm for multi-modal and HS data processing. Although these methods mentioned above might provide a feasible way for the CML-related issues, yet the ability to extract the discriminative features remains limited. This possibly results from the lack of directly modeling the latent subspace and label information. 

To facilitate the improvement of feature discrimination, we propose to simultaneously learn the shared subspace and regress the labels from the learned subspace in a joint fashion. Inspired by MA, we also enforce a graph-based alignment constraint on the multi-modal data, aiming at a more compact subspace learning. In fact, the proposed method in this letter is an extended version of common subspace learning (CoSpace) presented in \cite{hong2019cospace}. The main differences lie in two aspects. On one hand, we emphatically analyze the effects of different regression strategies, such as ridge regression ($\ell_{2}$-penalty), sparse regression ($\ell_{1}$-penalty). On the other hand, we further investigate the potentials of CoSpace-based models while handling the heterogeneous data (e.g., MS-Lidar and HS).

\begin{figure*}[!t]
	  \centering
		\subfigure{
			\includegraphics[width=0.85\textwidth]{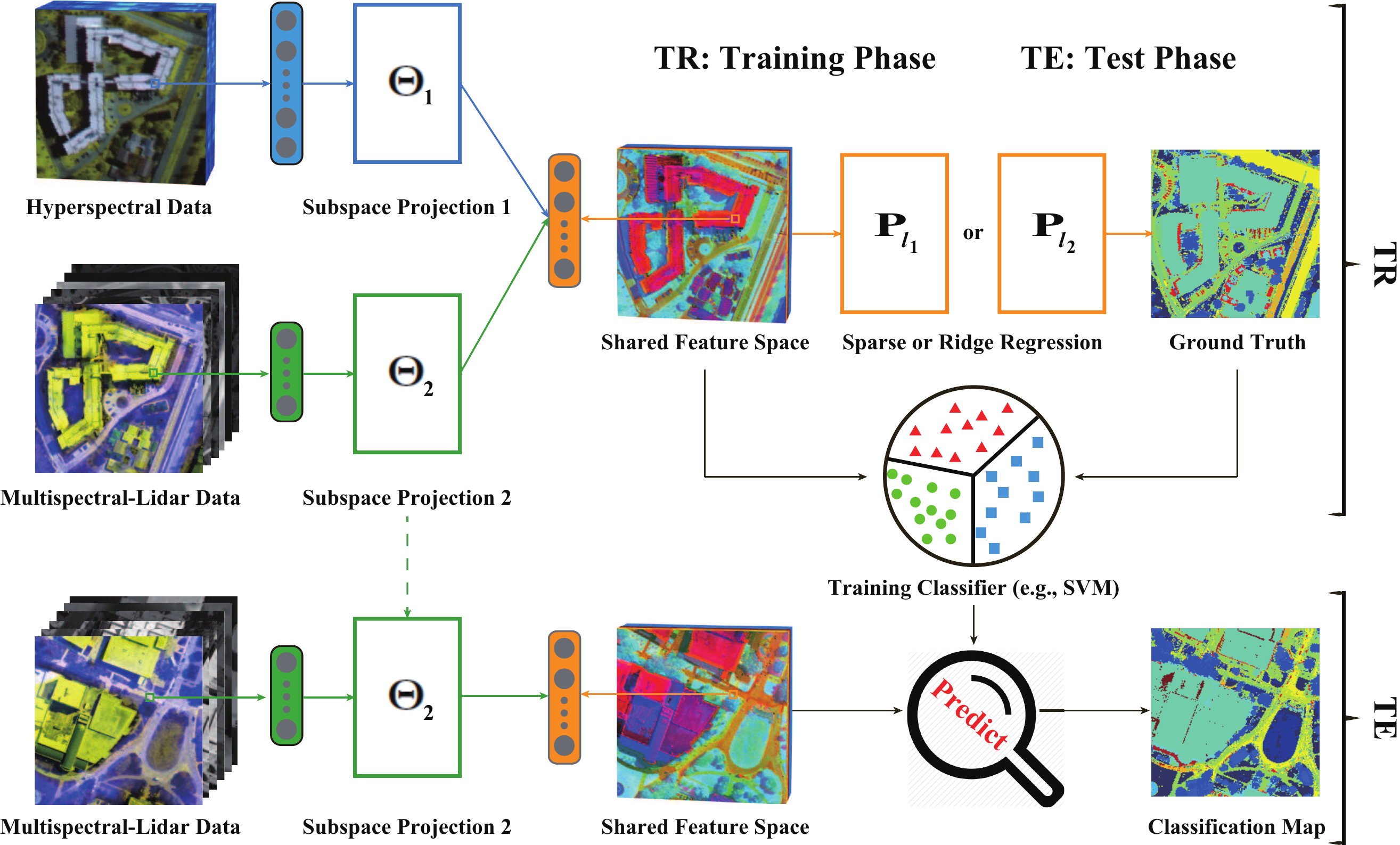}
		}
        \caption{An illustration of the proposed cross-modality feature learning. The CML problem specifically refers to model learning using multi-modalities in the training phase and testing the model only using one of multi-modalities (please see the Section II.A: Brief Motivation for more details).}
\label{fig:workflow}
\end{figure*}
\section{Methodology}
Fig. \ref{fig:workflow} illustrates the proposed cross-modality feature learning framework. In the section, we start with a review of the existing CoSpace model, and then discuss and analyze the potentials of using sparse regression in CoSpace. Finally, an optimizer based on alternating direction method of multipliers (ADMM) is briefly introduced to solve the extended CoSpace.

\subsection{Brief Motivation}
Although operational optical satellites, e.g., Sentinel-2 and Landsat-8, enable the MS data openly and largely available on a global scale, the MS data fail to distinguish similar classes due to its few spectral bands. Rather, the HSI is acquired with rich spectral information, enabling identification of the materials more easily and accurately, but its space coverage is far narrower than that of MS data. This naturally motivates us to investigate a general but interesting question -- \emph{can a limited amount of HS data partially overlapping with MS data improve the classification performance of the extra large-scale and non-overlapped MS data?} This is a typical CML-based problem setting.
\subsection{Review of CoSpace}
For this purpose, we proposed a feasible solution in \cite{hong2019cospace}, namely $\ell_{2}$-CoSpace. The proposed method, supporting an multi-modal input in the training phase, aims at learning a common subspace from multi-modalities, where the learned features are expected to be discirminative by fusing the different modality-specific information as much as possible. Theoretically speaking, through the shared feature space, the different modalities can be arbitrarily translated each other. We also connected the learned features and the label information by means of regression techniques for a more discirminative representation. Moreover, simultaneously considering the above strategies leads to the following joint model.

Given two modalities $\mathbf{X}_{1} \in\mathbb{R}^{d_{1}\times N}$ and $\mathbf{X}_{2} \in\mathbb{R}^{d_{2}\times N}$, namely HS data with $d_{1}$ bands by $N$ pixels and MS-Lidar data with $d_{2}$ bands by $N$ pixels in our case, the CoSpace can be modeled by optimizing the following objective function.
\begin{equation}
\label{eq1}
\mathop{\min}_{\mathbf{P},\mathbf{\Theta}}\left\{
\begin{aligned}
\frac{1}{2}\norm{\mathbf{\widetilde{Y}}-\mathbf{P}\mathbf{\Theta}\mathbf{\widetilde{X}}&}_{\F}^{2}+\frac{\alpha}{2}\norm{\mathbf{P}}_{\F}^{2}+\frac{\beta}{2}\tr(\mathbf{\Theta}\mathbf{\widetilde{X}}\mathbf{L}(\mathbf{\Theta}\mathbf{\widetilde{X}})^{\T})\\
&\mathrm{s.t.} \quad \mathbf{\Theta}\mathbf{\Theta}^{\T}=\mathbf{I}
\end{aligned}
\right\},
\end{equation}
where $\mathbf{Y} \in\mathbb{R}^{L\times N}$ denotes the one-hot encoded label matrix and $\mathbf{\widetilde{Y}}$ is defined as $\lbrack \mathbf{Y}, \mathbf{Y}\rbrack \in\mathbb{R}^{L\times 2N}$. $\widetilde{\mathbf{X}}=
\begin{bmatrix}
     \mathbf{X}_{1} & \mathbf{0} \\
     \mathbf{0} & \mathbf{X}_{2}
\end{bmatrix} \in\mathbb{R}^{(d_{1}+d_{2})\times 2N}$ and $\mathbf{\Theta}=\lbrack \mathbf{\Theta}_{1}, \mathbf{\Theta}_{2}\rbrack \in\mathbb{R}^{d\times (d_{1}+d_{2})}$ represents the subspace projection with respect to $\mathbf{X}_{1}$ and $\mathbf{X}_{2}$. $d$ is the dimension of the learned subspace. The variable $\mathbf{P} \in\mathbb{R}^{L \times d}$ is the regression matrix regularized by $\ell_{2}$-norm, which connects the latent subspace and label information for a discriminative feature representation. Moreover, $\mathbf{L}=\mathbf{D}-\mathbf{W}\in\mathbb{R}^{2N\times 2N}$ is defined as a joint Laplacian matrix, and $\mathbf{W}$ is the corresponding adjacency matrix formulated as follows
\begin{equation}
\label{eq2}
  \mathbf{W}^{i,j}=
    \begin{cases}
      \begin{aligned}
      1/N_k, \;\; & \text{if \(\mathbf{X}^{i}\) and \(\mathbf{X}^{j}\) are the \(k\)-th class;}\\
      0, \;\; & \text{otherwise,}
      \end{aligned}
    \end{cases}
\end{equation}
and then $\mathbf{D}$ is computed by $\mathbf{D}^{ii}=\sum_{i\neq j}\mathbf{W}^{i,j}$. 

For the model solution in Eq. (\ref{eq1}), we adopt an iterative alternating optimization strategy \cite{hong2019augmented} to convert the nonconvexity of Eq. (\ref{eq1}) to the convex subproblems of each variable $\mathbf{P}$ and $\mathbf{\Theta}$. The optimization subproblem with respect to the variable $\mathbf{P}$ is a typical least-squares problem with Tikhonov regularization, which has an analytical solution of $ \mathbf{P} = (\widetilde{\mathbf{Y}}\mathbf{Q}^{\T})(\mathbf{Q}\mathbf{Q}^{\T}+\alpha\mathbf{I})^{-1}$, where $\mathbf{Q}=\mathbf{\Theta}\widetilde{\mathbf{X}}$. For the optimization problem of $\mathbf{\Theta}$, it can be effectively and efficiently solved by ADMM.  Please refer to \cite{hong2019cospace} for more details.

\subsection{Sparse Regression based CoSpace ($\ell_{1}$-CoSpace)}
To further improve the CoSpace's representation ability, we propose to model the sparsity-promoting regression matrix, yielding a $\ell_{1}$-CoSpace. Unlike the original CoSpace model with $\ell_{2}$-regularized ridge regression ($\ell_{2}$-CoSpace for short), $\ell_{1}$-CoSpace learns a sparse regression matrix to connect the latent subspace and label space. More specifically, the advantages of $\ell_{1}$-norm over $\ell_{2}$-norm can be summarized as follows:
\begin{itemize}
    \item Compared to the $\ell_{2}$-norm, it is well known that the $\ell_{1}$-norm plays a role of feature selection, which makes the learned features more robust and further enhances the model's generalization ability.
    \item As introduced in \cite{cai2007spectral}, the sparsity-based learning or regression technique is capable of better interpreting the intrinsic structure of the data (or feature) space. This might effectively excavate or discover the underlying correspondences between selected features and certain classes, thereby yielding more effective feature learning.
\end{itemize}
\begin{algorithm}[!t]
\label{alg1}
\caption{CoSpace-based solution}
\KwIn{$\widetilde{\mathbf{Y}}$, $\widetilde{\mathbf{X}}$, $\mathbf{L}$, and parameters $\alpha$, $\beta$, $\maxIter$.}
\KwOut{$\mathbf{P}$, $\mathbf{\Theta}$}
 $t=1$, $\zeta=1e-4$;\\
{\textbf{Initializing} $\mathbf{P}$ and $\mathbf{\Theta}$}\\

  \While{not converged \rm{or} $t>\maxIter$}
 {
   Fix other variables to update $\mathbf{P}$

   Fix other variables to update $\mathbf{\Theta}$
   
   Compute the objective function value $E^{t+1}$ and check the convergence condition:
   \eIf{$|\frac{E^{t+1}-E^{t}}{E^{t}}|<\zeta$}
   {
     Stop iteration;
   }
   {
     $t\leftarrow t+1$;
   }
 }
\end{algorithm}
Accordingly, the resulting $\ell_{1}$-CoSpace can be formulated as
\begin{equation}
\label{eq4}
\mathop{\min}_{\mathbf{P},\mathbf{\Theta}}\left\{
\begin{aligned}
\frac{1}{2}\norm{\mathbf{\widetilde{Y}}-\mathbf{P}\mathbf{\Theta}\mathbf{\widetilde{X}}&}_{\F}^{2}+\alpha\norm{\mathbf{P}}_{1,1}+\frac{\beta}{2}\tr(\mathbf{\Theta}\mathbf{\widetilde{X}}\mathbf{L}(\mathbf{\Theta}\mathbf{\widetilde{X}})^{\T})\\
&\mathrm{s.t.} \quad \mathbf{\Theta}\mathbf{\Theta}^{\T}=\mathbf{I}
\end{aligned}
\right\},
\end{equation}
where $\norm{\textbf{P}}_{1,1}\equiv \sum\limits_{k=1}^N\norm{\textbf{p}_k}_1$ is used to approximate the sparsity.

Similarly to $\ell_{2}$-CoSpace, the problem (4) can be separated into two convex subproblems for the variables $\mathbf{P}$ and $\mathbf{\Theta}$, respectively. Moreover, the optimization problem of $\mathbf{P}$ can be quickly solved by the well-known \textit{soft threshold} operator \cite{hong2018sulora} under the ADMM framework, while the solution for the variable $\mathbf{\Theta}$ is same with that in $\ell_{2}$-CoSpace.  \textbf{Algorithm 1} details the specific solutions for the problem (1) or (4).

\section{Experiments}
To assess the performance of CoSpace-related methods (e.g., $\ell_{2}$-CoSpace, $\ell_{1}$-CoSpace) compared to several state-of-the-art baselines, we explore the classification as a potential application in terms of \textit{Overall Accuracy} (OA), \textit{Average Accuracy} (AA), and \textit{Kappa Coefficient} ($\kappa$). Two popular classifiers: linear support vector machines (LSVM) and canonical correlation forests (CCF) \cite{rainforth2015canonical}, are used in our case.
\subsection{Data Description}
\begin{figure}[!t]
	  \centering
		\subfigure[Hyperspectral data]{
			\includegraphics[width=0.45\textwidth]{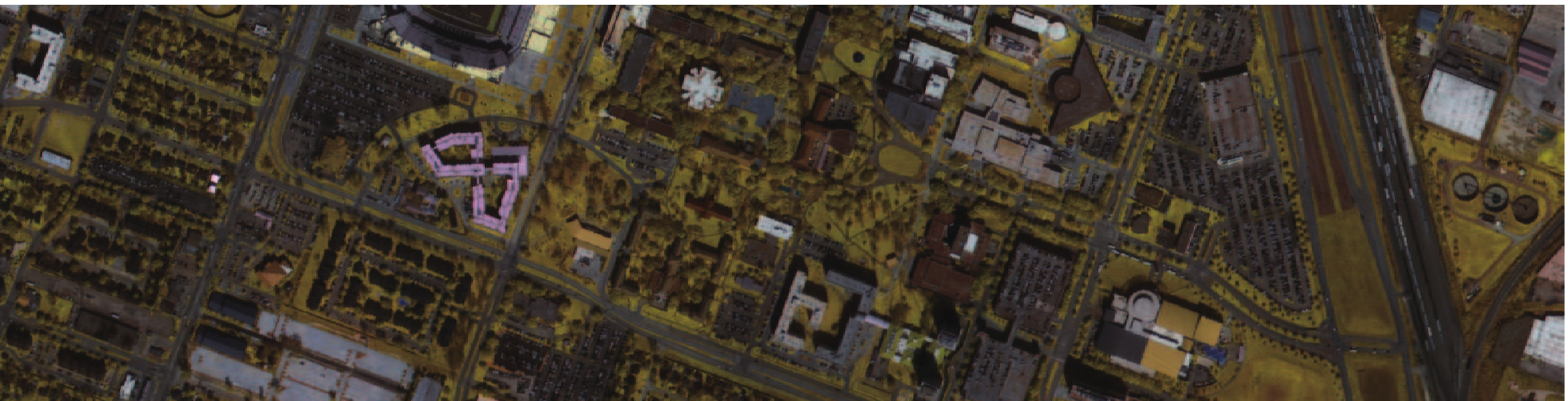}
		}
		\subfigure[Multispectral-Lidar data]{
			\includegraphics[width=0.45\textwidth]{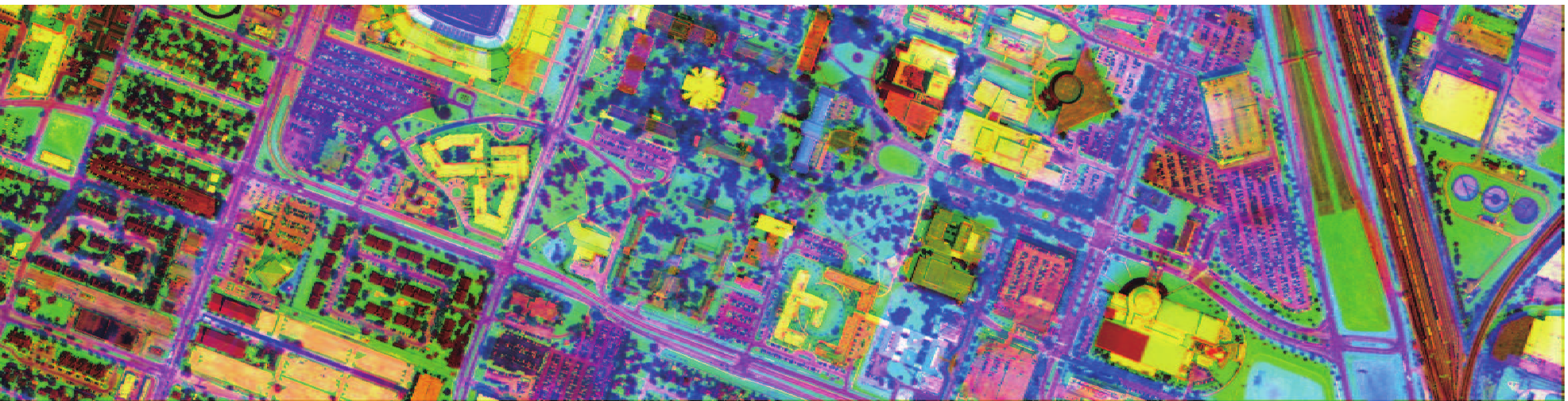}
		}

         \caption{False-color images of the used two data (HS and MS-Lidar data).}
\label{fig:FalseColor}
\end{figure}

We conducted the experiments on MS-Lidar and HS data provided by the 2018 IEEE GRSS data fusion contest (DFC2018) during the training phase \cite{le20182018}, where the MS-Lidar data and HSI were acquired by an Optech Titam MW (14SEN / CON340) with a Lidar sensor and an ITRES CASI 1500 sensor, respectively. The MS-Lidar data was collected from three different wavelengths (1550 nm, 1064 nm, and 532 nm) at a 50 cm ground sampling distance (GSD). It consists of $1202\times 4768$ pixels with nine bands (three downsampled RGB bands, three intensity bands, and 3 DEM bands). Note that these bands are stacked \cite{hong2015novel} as the model input. The corresponding HSI with the dimensions of $601\times 2384 \times 48$ covers the wavelength range from 380 nm to 1050 nm at a GSD of 1 m. The false-color images for the two used data are shown in Fig. \ref{fig:FalseColor}.
\begin{table*}[!t]
\centering
\caption{Quantitative performance comparisons of different algorithms in terms of OA, AA, and $\kappa$. The best one is shown in bold.}
\resizebox{\textwidth}{!}{
\begin{tabular}{p{43pt}<{\centering}||p{30pt}<{\centering}|p{30pt}<{\centering}||p{30pt}<{\centering}|p{30pt}<{\centering}||p{30pt}<{\centering}|p{30pt}<{\centering}||p{30pt}<{\centering}|p{30pt}<{\centering}||p{30pt}<{\centering}|p{30pt}<{\centering}||p{30pt}<{\centering}|p{30pt}<{\centering}}
\toprule[1.5pt]Algorithm&\multicolumn{2}{c||}{Baseline (\%) }&\multicolumn{2}{c||}{P-JDR (\%) }&\multicolumn{2}{c||}{L-USMA (\%) }&\multicolumn{2}{c||}{L-SMA (\%) }&\multicolumn{2}{c||}{$\ell_{2}$-CoSpace (\%)}&\multicolumn{2}{c}{$\ell_{1}$-CoSpace (\%)}\\
\hline \hline Parameter &\multicolumn{2}{c||}{\multirow{2}{*}{$(9,-,-,-,-)$}}&\multicolumn{2}{c||}{\multirow{2}{*}{$(20,-,-,-,-)$}}&\multicolumn{2}{c||}{\multirow{2}{*}{$(30,10,1,-,-)$}}&\multicolumn{2}{c||}{\multirow{2}{*}{$(20,-,-,-,-)$}}&\multicolumn{2}{c||}{\multirow{2}{*}{$(30,-,-,0.1,0.1)$}}&\multicolumn{2}{c}{\multirow{2}{*}{$(30,-,-,0.1,0.01)$}}\\
$(d, k, \sigma, \alpha, \beta)$&\multicolumn{2}{c||}{}&\multicolumn{2}{c||}{}&\multicolumn{2}{c||}{}&\multicolumn{2}{c||}{}&\multicolumn{2}{c||}{}\\
\hline \hline Classifier & LSVM & CCF & LSVM & CCF & LSVM & CCF & LSVM & CCF & LSVM & CCF & LSVM & CCF\\
\hline \hline 
OA&53.77&59.79&52.60&62.59&53.21&62.46&54.76&62.76&57.73&66.25&58.15&\bf68.17\\              
AA&48.26&47.94&47.17&53.91&49.03&54.00&52.67&56.45&53.17&60.78&56.43&\bf62.28\\
$\kappa$&0.4531&0.5094&0.4408&0.5426&0.4478&0.5420&0.4661&0.5481&0.4931&0.5892&0.5007&\bf0.6093\\
\hline \hline 
Class1&77.16&57.75&78.62&65.75&81.72&69.03&80.34&72.07&\bf83.41&71.60&80.97&76.95\\             
Class2&45.62&52.65&41.66&81.88&44.56&67.88&50.85&76.21&61.42&\bf87.16&47.62&85.79\\
Class3&\textbf{100.00}&\textbf{100.00}&\textbf{100.00}&\textbf{100.00}&\textbf{100.00}&\textbf{100.00}&\textbf{100.00}&\textbf{100.00}&\textbf{100.00}&\textbf{100.00}&94.17&94.17\\    Class4&87.56&91.12&88.33&88.70&91.84&91.54&\bf95.19&92.44&95.09&93.25&93.67&91.83\\            
Class5&68.00&48.13&59.01&63.41&62.28&71.88&85.93&79.75&\bf89.56&81.69&88.17&85.02\\       
Class6&7.47&0.00&11.11&0.76&2.12&0.10&\bf15.46&0.00&2.94&0.00&1.18&0.00\\
Class7&23.50&0.00&0.55&21.31&40.44&12.57&42.62&39.89&41.53&45.90&\bf54.64&48.63\\              
Class8&79.91&72.78&78.89&67.79&76.99&71.86&75.08&74.46&72.46&\bf83.78&79.02&80.27\\
Class9&67.20&77.07&65.53&77.50&65.38&77.14&65.90&74.97&72.37&77.87&69.97&\bf80.57\\              
Class10&11.24&\bf31.17&14.81&28.90&8.70&28.17&21.78&29.81&21.29&25.83&19.09&\bf31.17\\
Class11&26.58&41.45&24.94&45.19&24.66&49.37&28.81&47.20&29.34&52.85&44.25&\bf56.63\\
Class12&31.73&35.94&33.63&\bf40.54&31.43&36.94&27.63&36.04&21.32&35.44&36.14&36.64\\
Class13&31.10&33.26&28.82&34.58&39.63&36.87&27.19&38.75&20.72&39.10&31.42&\bf44.25\\              
Class14&46.90&30.49&41.60&30.10&48.08&33.16&48.03&45.74&\bf53.85&47.71&52.49&50.05\\
Class15&58.66&43.43&59.55&40.60&\bf60.25&40.78&57.42&38.40&55.57&53.04&54.96&38.71\\
Class16&26.16&20.81&16.18&32.02&17.80&27.63&28.20&21.70&46.45&28.60&\bf37.80&31.59\\
Class17&0.00&7.38&0.00&9.40&0.00&10.74&6.04&16.78&2.01&39.60&23.49&\bf59.06\\
Class18&55.69&48.03&68.91&78.77&60.09&80.11&71.82&82.63&67.39&86.67&84.90&\bf91.28\\
Class19&64.32&\bf88.09&76.49&88.35&68.10&87.05&73.75&83.88&71.36&87.14&85.05&88.01\\
Class20&56.46&79.18&54.77&82.70&56.53&\bf86.36&51.31&78.33&55.35&78.39&49.54&75.00\\
\bottomrule[1.5pt]
\end{tabular}}
\label{tab:DFC2018}
\end{table*}
\subsection{Experimental Configuration}
All models involved in performance comparison are trained on MS-Lidar and HS data and tested only in presence of MS-Lidar data to meet our CML's problem setting. To have pixel-to-pixel aligned on the two different modalities, we downsampled the MS-Lidar and ground truth to the HSI's spatial resolution using the nearest neighbor interpolation. Notably, in the used dataset, the sample distribution between the classes is extremely unbalanced. To provide a more reasonable and meaningful performance analysis. we evenly select 200 samples from each class for training\footnote{To meet the CML's problem setting and investigate the transferability of the proposed method, the training samples are collected in a limited region.} and the rest samples for testing. Furthermore, ten replications were performed for the selection of training and test samples and their averaged results are finally used for the quantitative evaluation.

We highlight the effectiveness of the proposed methods ($\ell_{2}$-CoSpace and $\ell_{1}$-CoSpace) in the CML-based task by quantitatively comparing with several state-of-the-art baselines, including the original MS-Lidar data (Baseline), PCA-based joint dimensionality reduction (P-JDR) \cite{hong2019cospace}, locality preserving projection (LPP)-based unsupervised MA (L-USMA) \cite{he2004LPP}, and LPP-based supervised MA (L-SMA) \cite{hong2019learnable}.
 
For a fair comparison, we aim at maximizing the classification performance by selecting the optimal parameters of different algorithms using 10-fold cross-validation (CV) on the training data. There parameters are feature dimension ($d$), regularization parameters ($\alpha,\beta$) in CoSpace, and the number of nearest neighbors ($k$) as well as the standard deviation of Gaussian kernel function ($\sigma$). More specifically, we select the parameters $d$ and $k$ ranging from $\{10, 20,...,50\}$, while the optimal $\sigma$, $\alpha$, and $\beta$ can be found from $\{10^{-2}, 10^{-1}, 10^{0}, 10^{1}, 10^{2}\}$. Please note that the CV is a widely-used and very effective strategy in the machine learning community to determine the model parameters, as long as the training samples with labels are given.

\subsection{Results and Analysis}
\begin{figure*}[!t]
	  \centering
	  \includegraphics[width=0.835\textwidth]{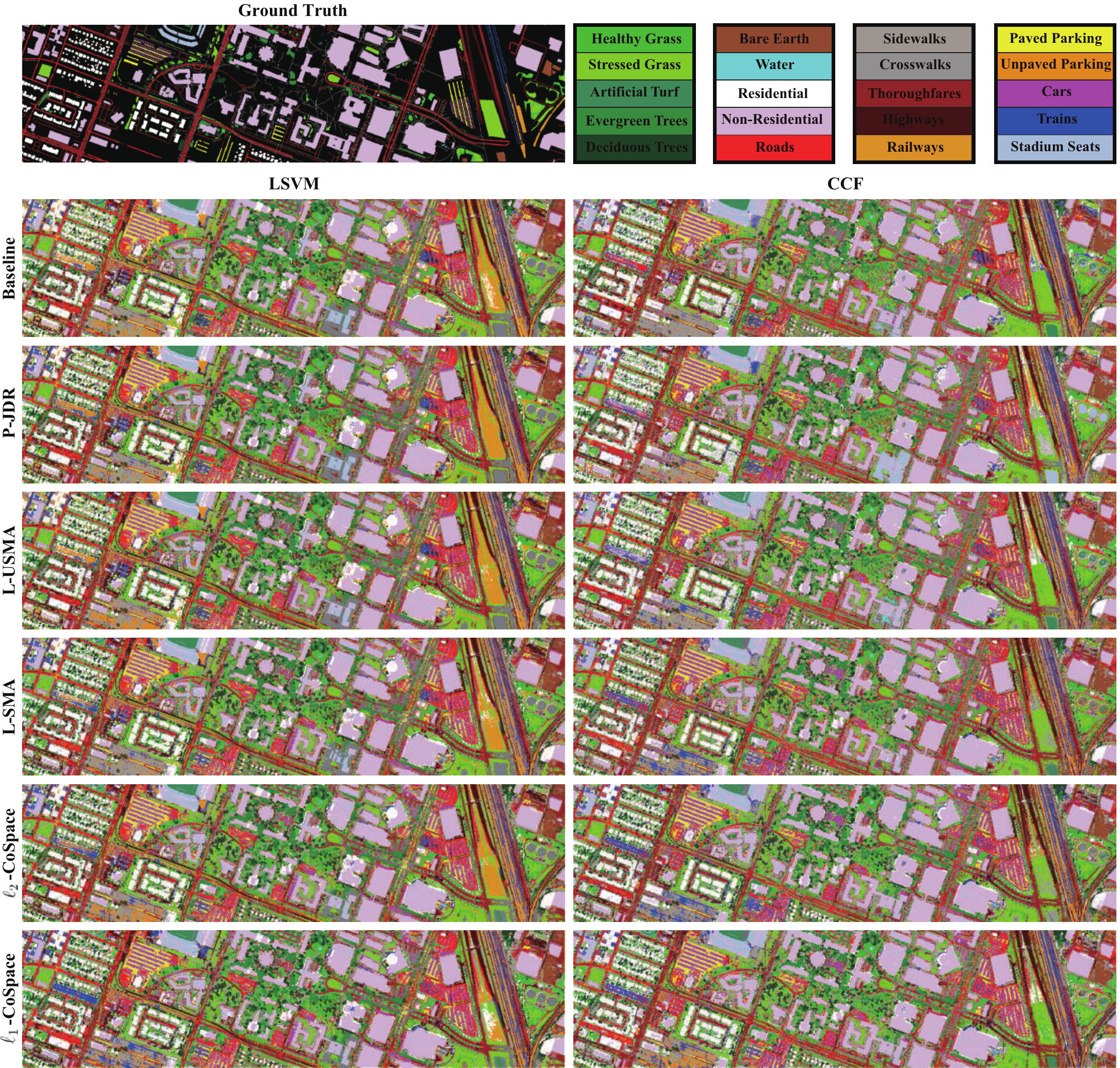}
         \caption{Classification maps predicted by different methods under the two classifiers (LSVM and CCF).}
\label{fig:CM}
\end{figure*}

The classification maps of different algorithms using LSVM and CCF classifiers are visualized in Fig. \ref{fig:CM}, and correspondingly the quantitative results for those compared methods in terms of OA, AA, and $\kappa$ are listed in  Table \ref{tab:DFC2018} where the parameters are experimentally determined by the 10-fold cross-validation on the training set. 

The baseline yields poor classification performance, due to the limitation of the feature representation ability. By jointly embedding MS-Lidar and HS data, P-JDR tends to obtain a higher classification accuracy than the baseline, particularly using CCF classifier (nearly 3\% improvement). Owing to fully considering the local topological structure of the input, L-USMA performs better than baseline and P-JDR. Similarly, L-SMA constructs an LDA-like graph based on the available labels, achieving better performance than L-USMA. Different with MA-based approaches (e.g., L-USMA, L-SMA) that attempt to directly find an aligned latent space from different modalities, the CoSpace-based models aim at jointly learning a latent subspace and a regression matrix bridging the learned subspace with labels. This might make the learned features more discriminative, thereby yielding the best classification performance. Remarkably still, we found that there is a further improvement in $\ell_{1}$-CoSpace over $\ell_{2}$-CoSpace. The possible reason for that is the use of sparse regression matrix, which is easy to implement the sparse-promoting structural learning.

With LSVM classifier, $\ell_{1}$-CoSpace improves the OAs of 4.35\%, 5.55\%, 4.94\%, 3.39\%, and 0.42\%, respectively, compared to baseline, P-JDR, L-USMA, L-SMA, and even $\ell_{2}$-CoSpace. While using CCF classifier, $\ell_{1}$-CoSpace obviously increases by 8.38\%, 5.58\%, 5.71\%, 5.41\%, and 1.92\% for the above five methods in terms of OA. As expected, the similar trends in AA and $\kappa$ can be also observed in Table \ref{tab:DFC2018}. 
\section{Conclusion}
In this letter, we investigate a CML-related problem by using the heterogeneous RS data (MS-Lidar and HS data). Concretely, we propose a novel joint sparse subspace learning ($\ell_{1}$-CoSpace) model, which is an improved version of CoSpace, by simultaneously learning a shared feature subspace and a sparse regression matrix. Benefiting from sparse modeling, the proposed $\ell_{1}$-CoSpace can interpret and mine the intrinsic structure of the data more effectively, resulting in a further performance improvement. 
\section*{Acknowledgment}
The authors would like to thank the Hyperspectral Image Analysis Laboratory at the University of Houston for acquiring and providing the data used in this study, and the IEEE GRSS Image Analysis and Data Fusion Technical Committee for organizing the 2018 IEEE GRSS Data Fusion Contest. 
\bibliographystyle{IEEEbib}
\bibliography{HDF_ref}

\end{document}